\documentclass[10pt,twocolumn,letterpaper]{article}

\usepackage{cvpr}
\usepackage{times}
\usepackage{epsfig}
\usepackage{graphicx}
\usepackage{amsmath}
\usepackage{amssymb}
\usepackage{pifont}

\usepackage[pagebackref=true,breaklinks=true,letterpaper=true,colorlinks,bookmarks=false]{hyperref}

\renewcommand{\vec}[1]{\boldsymbol{#1}}
\newcommand{\mat}[1]{\mathbf{#1}}

\newcommand{\template}[0]{\mat{T}}
\newcommand{\garment}[0]{\mat{G}}

\newcommand{\blendweights}[0]{\mat{W}}

\newcommand{\pose}[0]{\vec{\theta}}
\newcommand{\shape}[0]{\vec{\beta}}
\newcommand{\style}[0]{\vec{\gamma}}

\newcommand{\offsets}[0]{\mathbf{D}}
\newcommand{\stype}[0]{\phi}

\newcommand{\smpl}[0]{M}
\newcommand{\posefun}[0]{T}
\newcommand{\blendfun}[0]{W}
\newcommand{\offsetfun}[0]{B}
\newcommand{\offsetsfun}[0]{D}
\newcommand{\jointfun}[0]{J}

\newcommand{\cmark}{\ding{51}}%
\newcommand{\xmark}{\ding{55}}%

\makeatletter
\newcommand{\putindeepbox}[2][0.7\baselineskip]{{%
    \setbox0=\hbox{#2}%
    \setbox0=\vbox{\noindent\hsize=\wd0\unhbox0}
    \@tempdima=\dp0
    \advance\@tempdima by \ht0
    \advance\@tempdima by -#1\relax
    \dp0=\@tempdima
    \ht0=#1\relax
    \box0
}}
\makeatother

\makeatletter
\newcommand{\thickhline}{%
    \noalign {\ifnum 0=`}\fi \hrule height 1pt
    \futurelet \reserved@a \@xhline
}
\makeatother

\cvprfinalcopy 

\ifcvprfinal\fi
\begin{document}

\title{TailorNet: Predicting Clothing in 3D as a Function of Human Pose, Shape and Garment Style }
\author{Chaitanya Patel\footnotemark[1] \qquad Zhouyingcheng Liao\footnotemark[1] \qquad Gerard Pons-Moll\\
Max Planck Institute for Informatics, Saarland Informatics Campus, Germany\\
{\tt\small \{cpatel, zliao, gpons\}@mpi-inf.mpg.de}}

\makeatletter
\let\@oldmaketitle\@maketitle
\renewcommand{\@maketitle}{
	\@oldmaketitle
	\centering
	\vspace{-5mm}
	\includegraphics[width=0.95\textwidth]{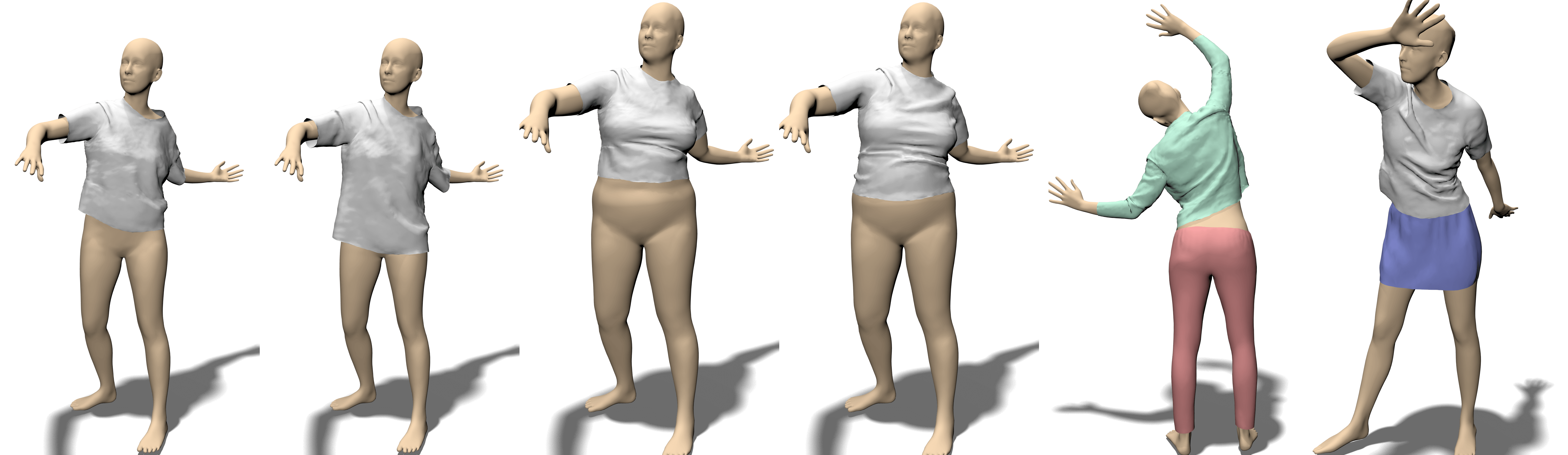}\\
	\refstepcounter{figure}\normalfont Figure~\thefigure: We present TailorNet, a model to estimate the clothing deformations with fine details from input body shape, body pose and garment style.
	From the left: the first two avatars show two different styles on the same shape, the following two show the same two styles on another shape, and the last two avatars illustrate that our method works for different garments.
	
	\label{fig:teaser}
	\vspace{5mm}
}
\makeatother

\maketitle

\renewcommand{\thefootnote}{\fnsymbol{footnote}}
\footnotetext[1]{Equal contribution.}
\begin{abstract}
In this paper, we present TailorNet, a neural model which  predicts clothing deformation in 3D as a function of three factors: pose, shape and style (garment geometry), while retaining wrinkle detail.  
This goes beyond prior models, which are either specific to one style and shape, or generalize to different shapes producing smooth results, despite being style specific.
Our hypothesis is that (even non-linear) combinations of examples smooth out high frequency components such as fine-wrinkles, which makes learning the three factors jointly hard. 
At the heart of our technique is a decomposition of deformation into a high frequency and a low frequency component. 
While the low-frequency component is predicted from pose, shape and style parameters with an MLP, the high-frequency component is predicted with a mixture of shape-style specific pose models. 
The weights of the mixture are computed with a narrow bandwidth kernel to guarantee that only predictions with similar high-frequency patterns are combined. 
The style variation is obtained by computing, in a canonical pose, a subspace of deformation, which satisfies physical constraints such as inter-penetration, and draping on the body. 
TailorNet delivers 3D garments which retain the wrinkles from the physics based simulations (PBS) it is learned from, while running more than $1000$ times faster.
In contrast to classical PBS, TailorNet is easy to use and fully differentiable, which is crucial for computer vision and learning algorithms.
Several experiments demonstrate TailorNet produces more realistic results than prior work, and even generates temporally coherent deformations on sequences of the AMASS~\cite{mahmood2019amass} dataset, despite being trained on static poses from a different dataset.
To stimulate further research in this direction, we will make a dataset consisting of 55800 frames, as well as our model publicly available at \href{https://virtualhumans.mpi-inf.mpg.de/tailornet/}{https://virtualhumans.mpi-inf.mpg.de/tailornet/}.
\end{abstract}

\section{Introduction}
\label{sec:introduction}
Animating digital humans in clothing has numerous applications in 3D content production, games, entertainment and virtual try on.
The predominant approach is still physics based simulation (PBS).
However, the typical PBS pipeline requires editing the garment shape in 2D with patterns, manually placing it on the digital character and fine tuning parameters to achieve desired results, which is laborious, time consuming and requires expert knowledge. Moreover, high quality PBS methods are computationally expensive, complex to implement and to control, and are not trivially differentiable. 
For computer vision tasks, generative models of clothed humans need to be differentiable, easy to deploy, and should be easy to integrate within CNNs, and to fit them to image and video data. 

In order to make animation easy, several works learn efficient approximate models from PBS complied off-line.
At least \emph{three factors} influence clothing deformation: body pose, shape and garment style (by style we mean the garment geometry).  
Existing methods either model deformation due to pose~\cite{Aguiar10,lahner2018deepwrinkles} for a fixed shape,
shape and pose~\cite{Casas19,Garnet19} for a fixed style,
or style~\cite{Mitra18} for a fixed pose.
The aforementioned methods inspire our work; however, they do not model the effects of pose, shape and style jointly, even though they are intertwined. Different garments deform differently as a function of pose and shape, and garment specific models~\cite{Casas19,lahner2018deepwrinkles,DRAPE12,Aguiar10} (1 style) have limited use. Furthermore, existing joint models of pose and shape~\cite{DRAPE12,Casas:2014,Garnet19} often produce over-smooth results (even for a fixed style), and they are not publicly available.
Consequently, none of existing approaches has materialized in a model which can be used to solve computer vision and graphics problems. 

What is lacking is a unified model capable of generating different garment styles, and animating them on any body shape in any pose, while retaining wrinkle detail.
To that end, we introduce \emph{TailorNet}, a mixture of Neural Networks (NN) learned from physics based simulations, which decomposes clothing deformations into style, shape and pose -- this effectively approximates the physical clothing deformation allowing intuitive control of synthesized animations.  
In the same spirit as SMPL~\cite{Loper2015SMPL} for bodies, TailorNet learns deformations as a displacements to a garment template in a canonical pose, while the articulated motion is driven by skinning.  
TailorNet can either take a real garment as input, or generate it from scratch, and drape it on top of the SMPL body for any shape and pose. In contrast to~\cite{DRAPE12}, our model predicts how the garment would fit in reality, \eg, a medium size garment is predicted tight on a large body, and loose on a thin body. 

Learning TailorNet required addressing several technical challenges. To generate different garment styles in a static pose, we compute a PCA subspace using the publicly available digital wardrobe of \emph{real static garments}~\cite{bhatnagar2019mgn}. To generate more style variation while satisfying garment-human physical constraints, we sample from the PCA subspace, run PBS for each sample, and recompute PCA again, to obtain a \emph{static style subspace}.  Samples from this subspace produce variation in sleeve length, size and fit in a static pose. 
To learn deformation as a function of pose and shape, we generated a semi-real dataset by animating garments (real or samples from the static style subspace) using PBS on top of SMPL~\cite{Loper2015SMPL} body for static SMPL poses, and for different shapes. 


Our first observation is that, for a \emph{fixed style} (garment instance) and body shape, predicting high frequency clothing deformations as a function of pose is possible -- perhaps surprisingly, our experiments show that, for this task, a simple multi-layer perceptron model (MLP) performs as well as or better than Graph Neural Networks~\cite{litany2017deformable,kolotouros2019cmr} and Image-decoder on a UV-space~\cite{lahner2018deepwrinkles}.
In stark contrast, straightforward prediction of deformation as a function of style, shape and pose results in overly smooth un-realistic results. 
We hypothesize that any attempt to combine training examples smoothes out \emph{high frequency} components, which explains why previous models~\cite{Casas19,DRAPE12,Garnet19}, even for a single style, lack fine scale wrinkles and folds. 

These key observations motivate the design of TailorNet: we predict the clothing low frequency geometry with a simple MLP. High frequency geometry is predicted with a mixture of high frequency style-shape specific models, where each specific model consists of a MLP which predicts deformation as a function of pose, and the weights of the mixture are obtained using a kernel which evaluates similarity in style and shape. A kernel with a very narrow bandwidth, prevents smoothing out fine scale wrinkles. 
Several experiments demonstrate that our model generalizes well to novel poses, predicts garment fit dependent on body shape, retains wrinkle detail, can produce style variations for a garment category (\eg, for T-shirts it produces different sizes, sleeve lengths and fit type), and importantly is easy to implement and control. 
To summarize, the main contributions of our work are: 
\begin{itemize}
	\item The first joint model of clothing style, pose and shape variation, which is simple, easy to deploy and fully differentiable for easy integration with deep learning.
	\item A simple yet effective decomposition of mesh deformations into low and high-frequency components, which coupled with a mixture model, allows to retain high-frequency wrinkles.
	\item A comparison of different methods (MLP, Graph Neural Nets and image-to-image translation on UV-space) to predict pose dependent clothing deformation.
	\item To stimulate further research, we make available a Dataset of $20$ aligned real static garments, simulated in $1782$ poses, for $9$ body shapes, totaling $55800$ frames. 
\end{itemize}
Several experiments show that our model generalizes to completely new poses of the AMASS dataset (even though we did not use AMASS to train our model), 
and produces variations due to pose, shape and style, while being more detailed than previous style specific models~\cite{Casas19,Garnet19}.
Furthermore, despite being trained on static poses, TailorNet produces smooth continuous animations. 
\section{Related Work}
\label{sec:related}
There are two main approaches to animation of clothing: physics based simulation (PBS), and efficient data-driven models learned from offline PBS, or real captures. 
\paragraph{Physics Based Simulation (PBS).}
Super realistic animations require simulating millions of triangles~\cite{terzopoulos1987elastically,selle2009robust,jiang2017anisotropic}, which is computationally expensive.  
Many works focus on making simulation efficient~\cite{GHFBG07} by adding wrinkles to low-resolution simulations \cite{Gillette:SCA:2015,Kavan:2011,Kim:2013:NEP,Wang:2010:EBW}, or
using simpler mass-spring models \cite{provot1995deformation} and position based dynamics~\cite{muller2007PBD,muller2008hpbd} compromising accuracy and physical correctness for speed.
Tuning the simulation parameters is a tedious task that can take weeks. Hence, several authors attempted to infer physical parameters mechanically~\cite{Miguel:2012:DEC,Wang:SIGGRAPH:2011}, from multi-view video~\cite{Rosenhahn:2007,yang2018physics,Stoll:2010}, or from perceptual judgments of humans~\cite{Sigal:2015:PCS}, but they only work in controlled settings and still require running PBS. 
PBS approaches typically require designing garments in 2D, grading them, adjusting them to the 3D body, and fine tuning parameters, which can take hours if not weeks, even for trained experts. 

\paragraph{Data-driven cloth models.}
One way to achieve realism is to capture real clothing on humans from images~\cite{bhatnagar2019mgn,alldieck2019tex2shape,alldieck2019learning,alldieck2018video,habermann2019TOG}, dynamic scans~\cite{Pons-Moll:Siggraph2017,neophytou2014layered} or RGBD~\cite{SimulCap19,DoubleFusion2018} and re-target it to novel shapes, but this is limited to re-animating the same motion on a novel shape. While learning pose-dependent models from real captures~\cite{yang2018analyzing,lahner2018deepwrinkles,ma20autoenclother} is definitely an exciting route, accurately capturing sufficient real data is still a major challenge. Hence,
these models~\cite{yang2018analyzing,lahner2018deepwrinkles,ma20autoenclother} only demonstrate generalization to motions similar to the training data.  

Another option is to generate data with PBS, and learn efficient data-driven models. 
Early methods relied on linear auto-regression or efficient nearest neighbor search to predict pose~\cite{Aguiar10,Kim:2013:NEP,wang2010example}, or pose and shape dependent clothing deformation~\cite{DRAPE12}.
Recent ones are based on deep learning, and vary according to the factor modeled (style or pose and shape), the data representation, and architecture used.  
Style variation is predicted from a user sketch with a VAE~\cite{Mitra18} but the pose is fixed. For a single style, pose and shape variation is regressed with MLPs and RNNs~\cite{Casas19,yang2018analyzing}, or Graph-NNs that process body and garment~\cite{Garnet19}. Pose effects are predicted with a normal map~\cite{lahner2018deepwrinkles} or a displacement map~\cite{pixelbased2018} in UV-space. These models~\cite{Aguiar10,Garnet19,DRAPE12} tend to produce over-smooth results, with the exception of~\cite{lahner2018deepwrinkles}, but the model is trained for a single garment and subject, and as mentioned before generalization to in the wild motions (for example CMU~\cite{de2009guide}) is not demonstrated. Since there is no consensus on what representation and learning model is best suited for this task, for a fixed style and shape, we compare representations and architectures, and find that MLPs perform as well as more sophisticated Graph-NNs or image-to-image translation on UV-space. While we draw inspiration from previous works, unlike our model, none of them can jointly model style, pose and shape variation. 
\paragraph{Pixel based models.}
An alternative to 3D cloth modeling is to retrieve and warp images and videos~\cite{Xu:2011:VCC,casasEG2014}. Recent methods use deep learning to generate pixels~\cite{han2017viton,Dong_2019_ICCV,Dong_2019_Towards,Yu_2019_ICCV,zhao2018multi,zhu2017your,Hsiao_2019_ICCV,raj2018swapnet,wang2018toward,GANprada17}, often guided by 2D warps, or smooth 3D models~\cite{Lassner:GP:2017,zanfir2018human,weng2018photo,shysheya2019textured}, or learn to transfer texture from cloth images to 3D models~\cite{mir20pix2surf}.
They produce (at best) photo-realistic images in controlled settings, but do not capture the 3D shape of wrinkles, cannot easily control motion, view-point, illumination, shape and style. 
 
\section{Garment Model Aligned with SMPL}
\label{sec:garmentmodel}
\begin{figure*}[ht]
	\centering
	\includegraphics[width=0.90\textwidth, trim={0 2cm 0 1cm},clip]{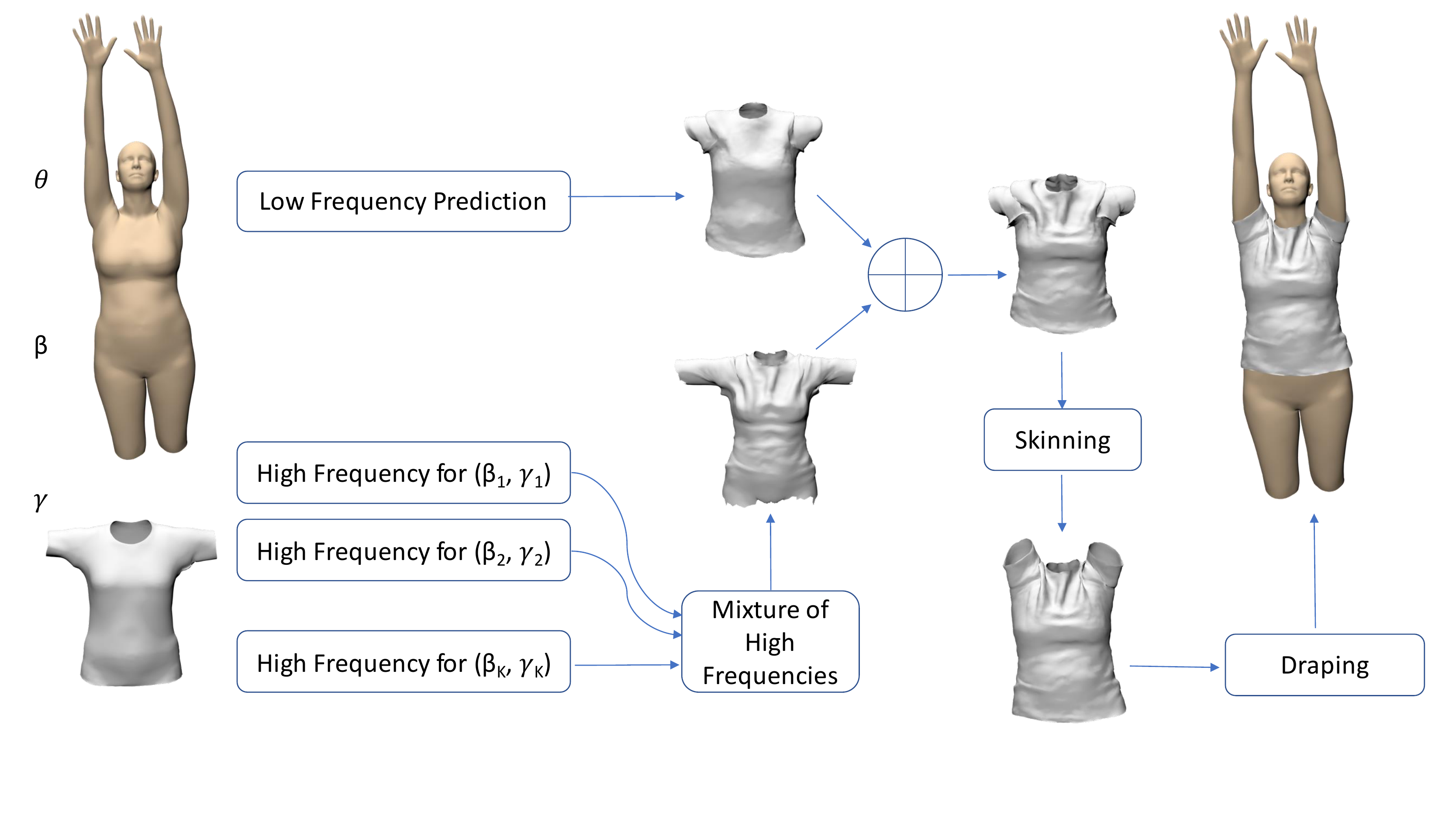}
	\caption{Overview of our model to predict the draped garment with style $\style$ on the body with pose $\pose$ and shape $\shape$. Low frequency of the deformations are predicted using a single model. High frequency of pose dependent deformations for $K$ prototype shape-style pairs are separately computed and mixed using a RBF kernel to get the final high frequency of the deformations. The low and high frequency predictions are added to get the unposed garment output, which is posed to using standard skinning to get the garment.}
	\label{fig:overview}
\end{figure*}

\begin{table*}
\centering
\begin{tabular}{| c c c c c c c |}
    \hline
    Method & Static/Dynamic & Pose & Shape & Style & Model & Dataset \\
    & & Variations & Variations & Variations & Public & Public \\
    \hline
    Santesteban \etal~\cite{Casas19} & Dynamic & \cmark & \cmark & \xmark & \xmark & \xmark \\
    Wang \etal~\cite{Mitra18} & Static & \xmark & \cmark & \cmark & \cmark & \cmark \\
    DeepWrinkles~\cite{lahner2018deepwrinkles} & Dynamic & \cmark & \cmark & \xmark & \xmark & \xmark \\
    DRAPE~\cite{DRAPE12} & Dynamic & \cmark & \cmark & \xmark & \xmark & \xmark \\
    GarNet~\cite{Garnet19} & Static & \cmark & \cmark & \xmark & \xmark & \cmark \\
    \hline
    Ours & Static & \cmark & \cmark & \cmark & \cmark & \cmark \\
    \hline
\end{tabular}
\caption{Comparison of our method with other works. Ours is the first method to model the garment deformations as a function of pose, shape and style. We also make our model and dataset public for further research.}
\label{table:allmethods}
\end{table*}
Our base garment template is aligned with SMPL~\cite{Loper2015SMPL} as done in \cite{Pons-Moll:Siggraph2017,bhatnagar2019mgn}.
SMPL represents the human body $\smpl(\cdot)$ as a parametric function of pose($\pose$) and shape($\shape$)
\begin{equation}
\label{eq:smplpose}
\smpl(\shape,\pose) = \blendfun(\posefun(\shape,\pose), \jointfun(\shape), \pose, \blendweights)
\end{equation}
\begin{equation}
\label{eq:smplshape}
\posefun(\shape,\pose) = \template + \offsetfun_s(\shape) + \offsetfun_p(\pose).
\end{equation}
composed of a linear function $\posefun(\shape,\pose)$ which adds displacements to base mesh vertices $\template\in \mathbb{R}^{n\times{3}}$ in a T-pose, followed by learned skinning $W(\cdot)$. 
Specifically, $\offsetfun_p(\cdot)$ adds pose-dependent deformations, and $\offsetfun_s(\cdot)$ adds shape dependent deformations. $\blendweights$ are the blend weights of a skeleton $\jointfun(\cdot)$. 

We obtain the garment template \textit{topology} as a submesh (with vertices $\template^G \in \mathbb{R}^{m\times{3}}$) of the SMPL template, with vertices $\template$.
Formally, the indicator matrix $\mat{I} \in \mathbb{Z}^{m\times{n}}$ evaluates to $\mathbf{I}_{i,j}=1$ if garment vertex $i \in \{1 \hdots m\}$ is associated with body shape vertex $j \in \{1 \hdots n\}$.
The particular garment \textit{style} draped over $\smpl(\shape,\pose)$ in a \emph{0-pose} is encoded as displacements $\offsets$ over the unposed body shape $T(\pose,\shape)$.
Since the garment is associated with the underlying SMPL body, previous works~\cite{bhatnagar2019mgn,Pons-Moll:Siggraph2017} deform every clothing vertex with its associated SMPL body vertex function.
For a given style $\offsets$, shape $\shape$ and pose $\pose$, they deform clothing using the un-posed SMPL function $T\pose,\shape)$
\begin{equation}
\label{eq:garment_shape}
\posefun^G(\shape,\pose,\offsets) = \mat{I}~\posefun(\shape,\pose)+\offsets,
\end{equation}
followed by the SMPL skinning function in Eq.~\ref{eq:smplpose} 
\begin{equation}
\label{eq:garment_pose}
    G(\shape,\pose,\offsets) = \blendfun(\posefun^G(\shape,\pose,\offsets), \jointfun(\shape), \pose, \blendweights).
\end{equation}
Since $\mathbf{D}$ is \emph{fixed}, this assumes that \emph{clothing deforms in the same way as the body}, which is a practical but clearly \emph{over-simplifying assumption}. 
Like previous work, we also decompose deformation as a non-rigid component (Eq.~\ref{eq:garment_shape}) and an articulated component (Eq.~\ref{eq:garment_shape}), 
but unlike previous work, we learn non-rigid deformation $\offsets$ as a function of pose, style and shape. That is, we learn true clothing variations as we explain in the next section.

\section{Method}
\label{sec:method}
In this section, we describe our decomposition of clothing as non-rigid deformation (due to pose, shape and style) and articulated deformation, which we refer to as \emph{un-posing} (Section~\ref{subsec:unposing}). 
The first component of our model is a subspace of garment styles which generates variation in A-pose (Section~\ref{method:genstyle}). 
As explained in the introduction, pose-models specific to a \emph{fixed shape and style} (explained in Section~\ref{method:singless}), do preserve high-frequencies, but models that combine different styles and shapes produce overly smooth results. Hence, at the heart of our technique is a model, which predicts low frequency with a straight-forward MLP, and high-frequency with a mixture model Sections~\ref{method:genstyle}--\ref{method:mixture}.
An overview of our method is shown in Fig.~\ref{fig:overview}.
\subsection{Un-posing Garment Deformation}
\label{subsec:unposing}
Given a set of \emph{simulated} garments $\garment$ for a given pose $\pose$ and shape $\shape$, we first 
disentangle non-rigid deformation from articulation by \emph{un-posing} -- we invert the skinning $\blendfun(\cdot)$ function
\begin{equation}
\label{eq:inverse_skinning}
\offsets = \blendfun^{-1}(\garment, \jointfun(\shape), \pose, \blendweights) - \mat{I}~\posefun(\shape,\pose),
\end{equation}
and subtract the body shape in a canonical pose, obtaining un-posed non-rigid deformation $\offsets$ as displacements from the body.
Since joints $J(\cdot)$ are known, computing $W^{-1}(\cdot)$ in Eq.~\ref{eq:inverse_skinning} entails un-posing every vertex $\hat{\mathbf{v}}_j = (\sum_k \mathbf{W}_{k,j} \mathbf{H}_k(\pose, J(\shape)))^{-1}\mathbf{v}_j$, where $\mathbf{H}_k(\pose, J(\shape)) \in SE(3)$ are the part transformation matrices, and $\hat{\mathbf{v}}_j \in \mathbb{R}^3$ is the un-posed vertex. 
Non-rigid deformation $\mathbf{D}$ in the unposed space is affected by body pose, shape and the garment style (size, sleeve length, fit).
Hence, we propose to learn deformation $\offsets$ as a function of shape $\shape$, pose $\pose$ and style $\style$, 
i.e. $\offsetsfun(\shape, \pose , \style): \mathbb{R}^{|\pose|} \times{\mathbb{R}^{|\shape|}}\times{\mathbb{R}^{|\style|}} \mapsto \mathbb{R}^{m\times{3}}$. The model should be realistic, easy to control and differentiable.  

\subsection{Generating Parametric Model of Style} \label{method:genstyle}
\begin{figure}[ht]
	\centering
	\includegraphics[width=0.4\textwidth, trim={0 0 0 1cm},clip]{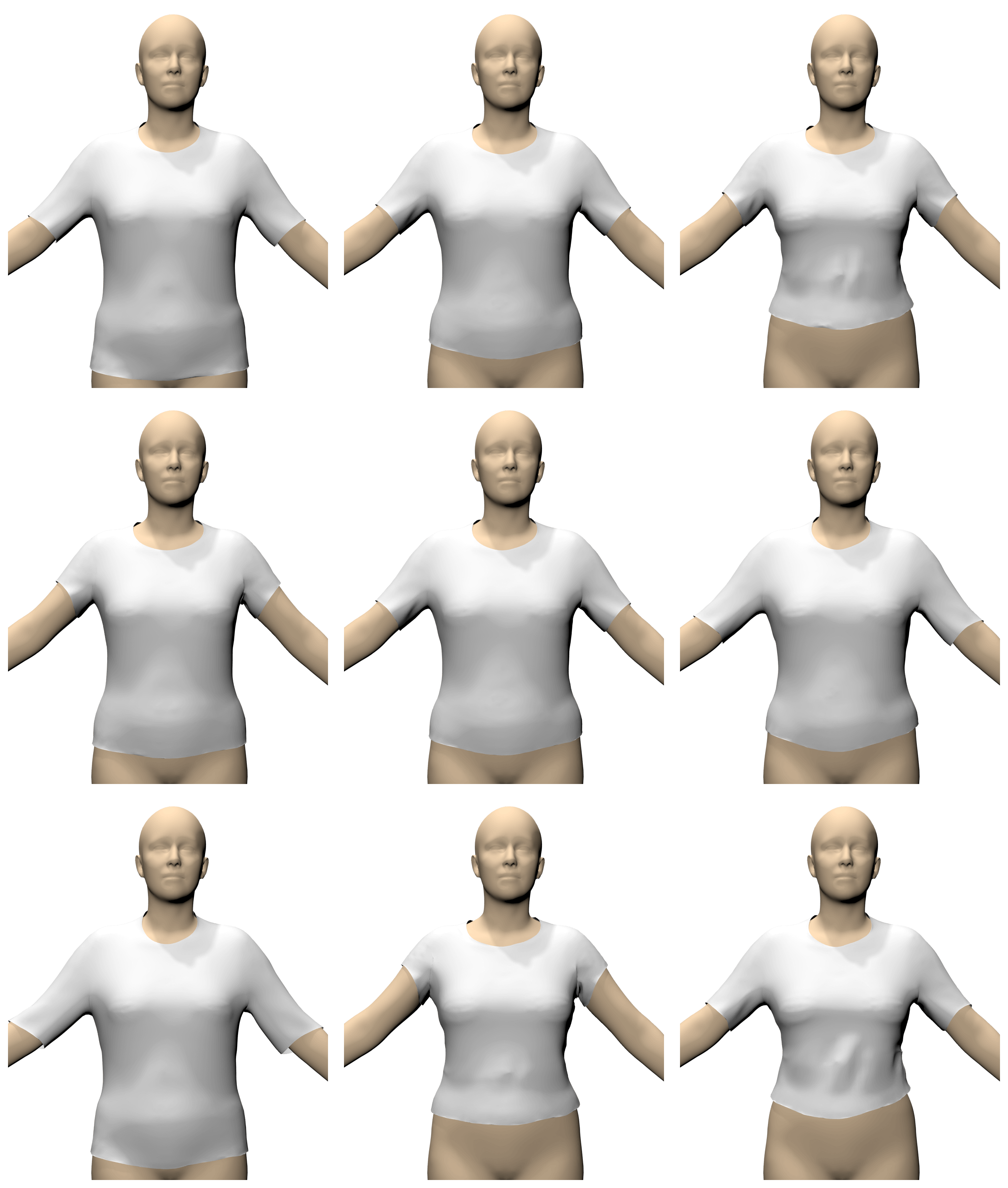}
	\caption{Overview of T-shirt style space. First component(top) and second component(middle) changes the overall size and the sleeve length respectively. Sampling from this style space generates a wide range of T-shirt styles (bottom). For the first two rows, corresponding components are $-1.5\sigma,0,1.5\sigma$ from left to right.}
	\label{fig:stylespace}
\end{figure}

We generate style variation $\style$ in A-pose by computing a PCA subspace using the public 3D garments of~\cite{bhatnagar2019mgn}. 
Although the method of Bhatnagar \etal~\cite{bhatnagar2019mgn} allows to transfer the 3D garment to a body shape in a canonical pose and shape ($\{G(\pose_0,\shape_0,\offsets_i)\}$), 
physical constraints might be violated -- garments are sometimes slightly flying on top of the body, and wrinkles do not correspond to an A-pose. 
In order to generate style variation while satisfying physics constraints, we alternate sampling form the PCA space and running PBS on the samples. 
We already find good results alternating PBS and PCA two times, to obtain a sub-space of style variation in a \emph{static A-pose} -- with PCA coefficients $\style$. 
Fig.~\ref{fig:stylespace} shows the parametric style space for t-shirt.
This is however, a model in a fixed A-pose and shape. Different garment \emph{styles will deform differently} as a function of \emph{pose} and \emph{shape}. 
In Section~\ref{method:singless} we explain our style-shape specific model of pose, and in Section~\ref{method:mixture} we describe our joint mixture model for varied shapes and styles.
\subsection{Single Style-Shape Model} \label{method:singless}
For a fixed garment style $\style$ and a fixed body shape $\shape$ pair, denoted as $\stype = (\style, \shape) \in \mathcal{X}$, we take paired simulated data of body poses and garment displacements $\{(\pose_i, \offsets^{\stype}_i)\}$ ,computed according to Eq.~\ref{eq:inverse_skinning}, and train a model $D_{\stype} : \mathbb{R}^{|\pose|} \xrightarrow{} \mathbb{R}^{m \times 3} $ to predict pose dependent displacements.
In particular, we train $D_{\stype}$ using a multi-layer perceptron(MLP) to minimize the L1-loss between the predicted displacements $D_{\stype}(\pose_i)$ and the ground-truth displacements $\offsets^{\stype}_i$. We observe that $D_{\stype}$ predicts reasonable garment fit along with fine-scale wrinkles and generalizes well to unseen poses, but this model is \emph{specific to a particular shape and style.} 
\subsection{TailorNet} \label{method:mixturemodel}
For the sake of simplicity, it is tempting to directly regress all non-rigid clothing deformations as function of pose, shape and style jointly, \ie, learning $D(\pose,\stype): \mathbb{R}^{|\pose|}\times{\mathcal{X}} \mapsto \mathbb{R}^{m\times{3}}$ directly with an MLP. However, our experiments show that, while this produces accurate predictions quantitatively,
qualitatively results are overly \emph{smooth lacking realism}. We hypothesize that any attempt to combine geometry of many samples with varying high frequency (fine wrinkles) smoothes out the details.
This might be a potential explanation for the smooth results obtained in related works~\cite{Casas19, Garnet19,DRAPE12} -- even for single style models. 
Our idea is to decompose garment mesh vertices, in an unposed space $\hat{\mathbf{G}} = \blendfun^{-1}(\garment, \jointfun(\shape), \pose, \blendweights)$, into a smooth low-frequency shape $\hat{\garment}^{LF}$, and a high frequency shape $\hat{\garment}^{HF}$ with diffusion flow. Let $f(\mathbf{x},t):\mathcal{G}\mapsto \mathbb{R}$ be a function on the garment surface, then it is smoothed with the diffusion equation:
\begin{equation}
\frac{\partial{f(\mathbf{x},t)}} {\partial t}= \lambda \Delta f(\mathbf{x},t)
\end{equation}
which states that the function changes over time by a scalar diffusion coefficient $\lambda$ times its spatial Laplacian $\Delta f$. 
In order to smooth mesh geometry, we apply the diffusion equation to the vertex coordinates $\mathbf{g}_i \in \hat{\mathbf{G}}$ (which are interpreted as a discretized function on the surface)
\begin{equation}
\mathbf{g}_i = \mathbf{g}_i + \lambda \Delta \mathbf{g}_i,
\label{eq:laplacian_smoothing}
\end{equation}
where $\Delta \mathbf{g}_i$ is the discrete Laplace-Beltrami operator applied at vertex $\mathbf{g}_i$, and $\lambda$ and the number of iterations control the level of smoothing. 
Eq.~\ref{eq:laplacian_smoothing} is also known as Laplacian smoothing~\cite{desbrun1999implicit,PGM}. We use $\lambda=0.15$ and 80 iterations, to obtain a smooth low-frequency mesh $\hat{\garment}^{LF}$ and a high-frequency residual $\hat{\garment}^{HF} = \hat{\garment}-\hat{\garment}^{LF}$. 
We then subtract the body shape $\mathbf{I} \posefun(\shape,\pose)$ as in Eq.~\ref{eq:inverse_skinning}, and predict displacement components separately as
 \begin{equation}
D(\pose,\stype) = D^{LF}(\pose,\stype) + \sum_{k=1}^K \Psi(\stype,\stype_k) D_{\stype,k}^{HF}(\pose),
 \end{equation}
where the low-frequency component is predicted with an MLP $D^{LF}(\pose,\stype)$ (smooth but accurate), whereas the high frequency component is predicted with a mixture of style-shape $\stype=(\style,\shape)$ specific models of pose $D_{\stype}^{HF}(\pose)$. As we show in the experiments, style-shape $\stype$ specific high frequency models retain details. We generalize to new shape-styles beyond the prototypes $D_{\stype_k}^{HF}(\pose)$ with a convex combination of specific models. The mixture weights are computed with a kernel $\Psi(\stype_1,\stype_2):\mathcal{X}\times{\mathcal{X}} \mapsto \mathbb{R}$ with a narrow bandwidth to \emph{combine only similar wrinkle patterns} 
\begin{equation}
\Psi(\stype,\stype_k) = \exp{\left(-\frac{\mathrm{dist}\left( g(\stype), g(\stype_k) \right)}{\sigma_x} \right)},
\end{equation}
where $g(\stype): \mathcal{X} \mapsto \mathbb{R}^{m\times{3}}$ is a shallow MLP which maps from style-shape to the garment displacements $\mathbf{D}$ in a canonical A-pose. 
Ideally, the kernel should measure similarity in style and shape, but this would require simulating training data for every possible pose-shape-style combination, which is inefficient and resource intensive. 
Our key simplifying assumption is that two garments on two different people will deform similarly if their displacements $g(\stype) = \mathbf{D}$ to their respective bodies is similar -- this measures clothing fit similarity. 
 While this is an approximation, it works well in practice. 

The bandwidth $\sigma_x$ is a free-parameter of our model, allowing to generate varying high-frequency detail. 
We find qualitatively good results by keeping  $\sigma_x$ small in order to combine only nearby samples.
We postprocess the output to remove the garment intersections with the body.
\label{method:mixture}

\paragraph{Choosing K Style-Shape Prototypes}
For each chosen style-shape pair, we simulate a wide variety of poses, which is time consuming. Hence, we find $K$ prototype shape-styles $\stype$ such that they cover the space of static displacements in A-pose $\offsetsfun(\shape, \pose_0, \style)$. Specifically, we want to choose $K$ style-shape pairs such that any other shape-styles can be approximated as a convex combination of prototypes with least error --  this is a non-convex problem with no global optimum guarantees. While iterative methods like K-SVD~\cite{aharon2006k} might yield better coverage, here we use a simple but effective greedy approach to choose good prototypes.

We take a dataset of $X$ garments $\{\offsetsfun(\pose_0, \stype)\}_{i=1...X}$ in different styles $\style_i$ draped on different body shapes $\shape_i$ in canonical pose.
We start with a pool of $9$ body shapes in one common style and try to fit each of the other style-shape pairs as a convex combination of this pool. 
Then we take the style-shape with the highest approximation error and add it to the pool. We repeat this process until we get the pool of $K$ style-shape pairs. We find all shape-styles can be well approximated when $K\geqslant20$. Thus, we use $K=20$ here. 

\section{Dataset}
\label{sec:dataset}

We learn our model from paired data of parameters and the garment which we obtained from simulation.
We use a commercial 3D cloth design and simulation tool Marvelous Designer~\cite{MarvellousDesigner} to simulate these garments. 
Instead of designing and positioning garments manually, we use the publicly available digital wardrobe~\cite{bhatnagar2019mgn} captured from the real world data which includes variations in styles, shapes and poses.
Although we describe our dataset generation with reference to T-shirt, method remains the same for all garments.

\subsection{Garments to Generate Styles}
We first retarget~\cite{Pons-Moll:Siggraph2017} the 3D garments from the digital wardrobe~\cite{bhatnagar2019mgn} to the canonical shape($\shape_0$) and slowly simulate them to canonical pose($\pose_0$).
We get the dataset $\{G(\shape_0, \pose_0, \offsets_i)\}$ of $43$ garments to learn style-space as in Section~\ref{method:genstyle}.
However, the learnt style-space may not be intuitive due to limited variation in the data, and may contain irregular patterns and distortions owing to the registration process.
So we sample 300 styles from this PCA model, simulate them again to get a larger dataset with less distortions.
With 2 iterations of PCA and simulation, we generate consistent and meaningful style variations.
We find first 2 PCA components enough to represent $\style$ parameters. See Figure \ref{fig:stylespace}.

\subsection{Shape, Pose and Style Variations}
We choose $9$ shapes manually as follows. 
We sample the first two shape components $\beta$ at $4$ equally spaced intervals, while leaving the other components to zero.
To these $4\times2=8$ shapes, we add canonical zero shape $\shape_0$. 
We choose $25$ styles in a similar way - by sampling the first two sytle $\style$ components. 

We simulate all combinations of shapes and styles in canonical pose, which results in $25\times9=225$ instances. 
We choose $20$ prototypes out of $225$ style-shape pairs as training style-shape pairs using the approach mentioned in Section~\ref{method:mixture}.
We randomly choose $20$ more style-shape pairs for testing.

\subsection{Simulation details} \label{dataset:Simulation}
For pose variations, we use $1782$ static SMPL poses, including a wide range of poses, including extreme ones.
For a given style-shape $\stype_k$ and poses $\{\pose_i\}$, we simulate them in one sequence. 
The body starts with canonical pose and greedily transitions to the nearest pose until all poses are traversed. 
To avoid dynamic effects (in this paper we are interested in quasi static deformation), 
we insert linearly interpolated intermediate poses, and let the garment relax for few frames. 
PBS simulation fails sometimes, and hence we remove frames with self-interpenetration from the dataset.

For training style-shape pairs, we simulate SMPL poses with interpolated poses.
For testing style-shape pairs, we simulate a subset of SMPL poses with a mix of seen and unseen poses.
Finally, we get 4 splits as follows:
(1) train-train : $20$ training style-shape pairs each with $2600$ training poses.
(2) train-test : $20$ training style-shape pairs each with $179$ unseen poses.
(3) test-train : $20$ testing style-shape pairs each with $35$ training poses.
(4) test-test : $20$ testing style-shape pairs each with $131$ unseen poses.

\section{Experiments}
\label{sec:experiments}
We evaluate our method on T-Shirt quantitatively and qualitatively and compare it to our baseline, and previous works. 
Our baseline and TailorNet use several MLPs - each of them has two hidden layers with ReLU activation and a dropout layer.
We arrive to optimal hyperparameters by tuning our baseline, and then keep them constant to train all other MLPs. 
See suppl. for more details.



\subsection{Results of Single Style-Shape Model}
We train single style-shape models $D_{\stype}^{HF}(\pose)$ on $K$ style-shape pairs separately. 
The mean per vertex test error of each model varies depending upon the difficulty of particular style-shape. 
The average error across all $K$ models is $8.04$ mm with maximum error of $14.50$ mm for a loose fitting, and minimum error of $6.56$ mm for a fit fitting.

For single style-shape model, in the initial stages of the work, we experimented with different modalities of 3D mesh learning: UV map, similar to~\cite{lahner2018deepwrinkles}, and graph convolutions (Graph CNN) on the garment mesh. 
For learning UV map, we train a decoder from the input pose parameters to UV map of the garment displacements. 
For Graph CNN~\cite{kipf2017semi} by putting the input $\pose$ on each node of the graph input. 
We report results on two style-shape pairs in Table~\ref{table:allpredictors}, which shows that a simple MLP is sufficient to learn pose dependent deformations for a fixed style-shape $\stype$.
\begin{figure}[ht]
	\centering
	\includegraphics[width=0.30\textwidth]{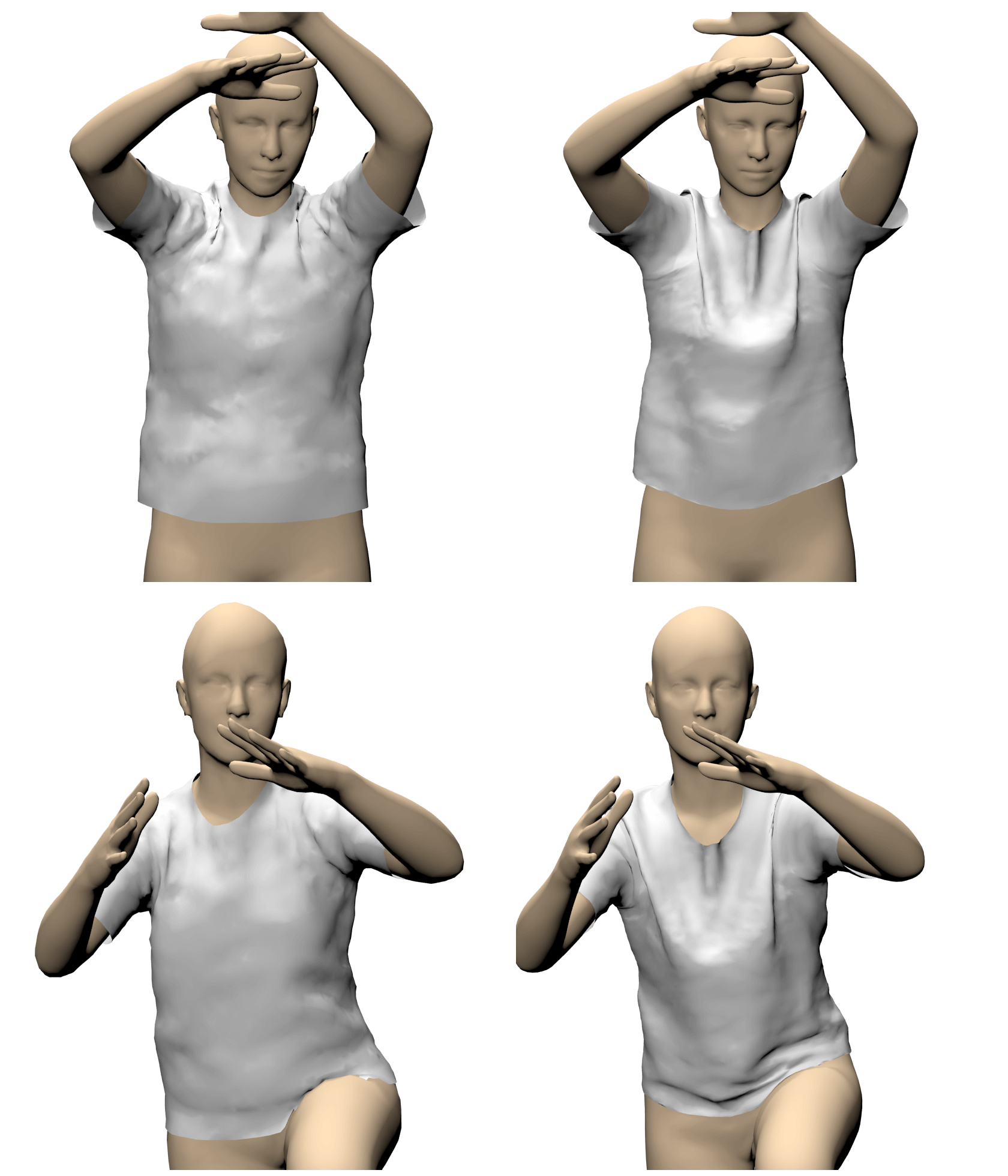}
	\caption{Results on unseen pose, shape and style. Prediction by \cite{Casas19} (left), our mixture model (right). Note that our mixture model is able to retain more meaningful folds and details on the garment.}
	\label{fig:compareEG}
\end{figure}

\begin{table}[ht]
\centering
\begin{tabular}{c|c c c}
    \thickhline
    Style-shape & MLP & UV Decoder & Graph CNN  \\
     \hline
    Loose-fit & 14.5 & 15.9 & 16.1 \\
    Tight-fit & 10.1 & 11.4 & 11.7 \\
    \thickhline
\end{tabular}
\caption{Mean per vertex error in mm for the pose dependent prediction by MLP, UV Decoder and Graph CNN for two style-shapes.}
\label{table:allpredictors}
\end{table}

\subsection{Results of TailorNet}
\begin{figure}[t]
	\centering
	\includegraphics[width=0.45\textwidth]{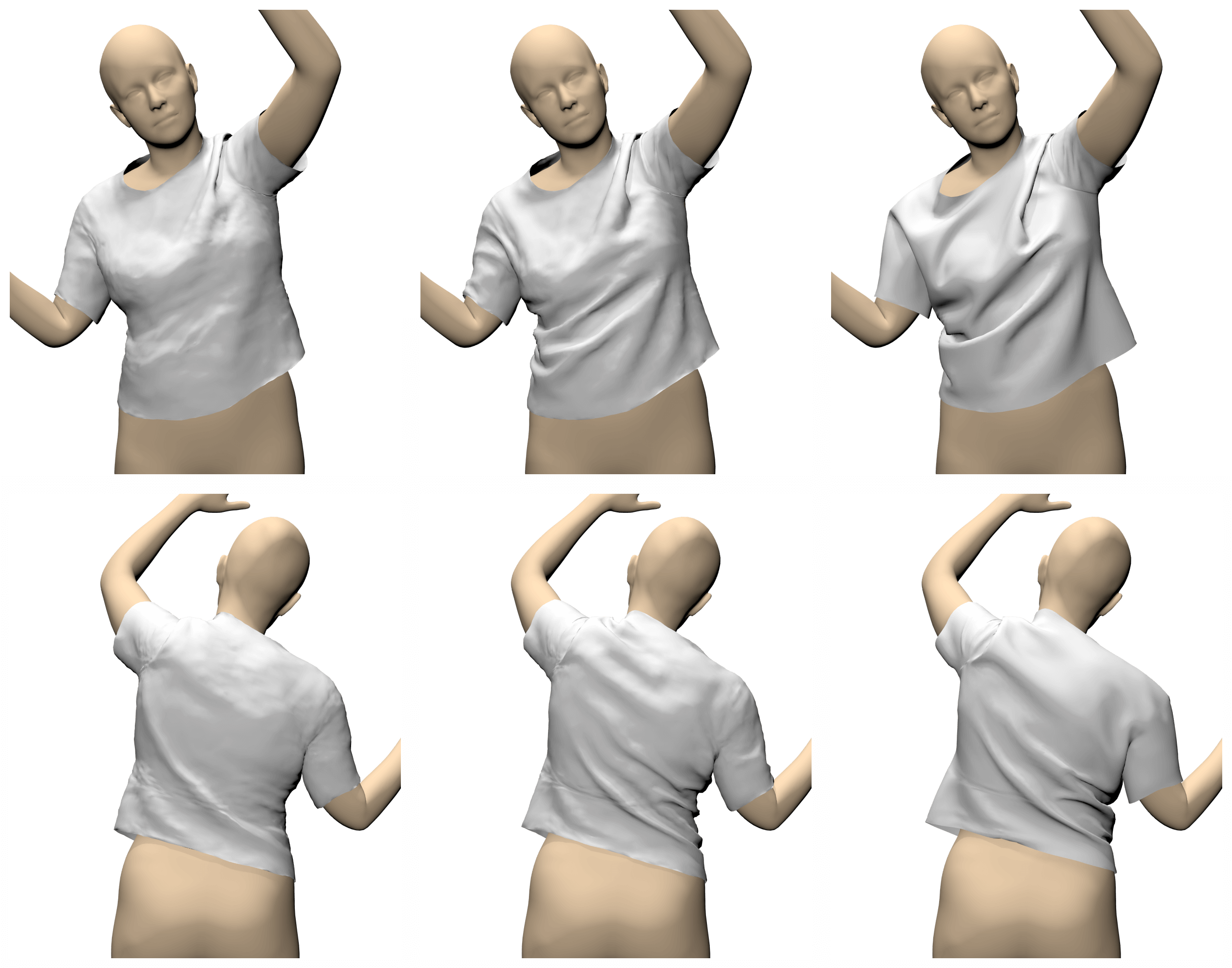}
	\caption{Baseline method (left) smooths out the fine details over the garment. TailorNet (middle) is able to retain as many wrinkles as PBS groundtruth (right).}
	\label{fig:compareBL}
\end{figure}
\begin{figure}[t]
	\centering
	\includegraphics[width=0.30\textwidth, trim={0 40cm 0 0cm},clip]{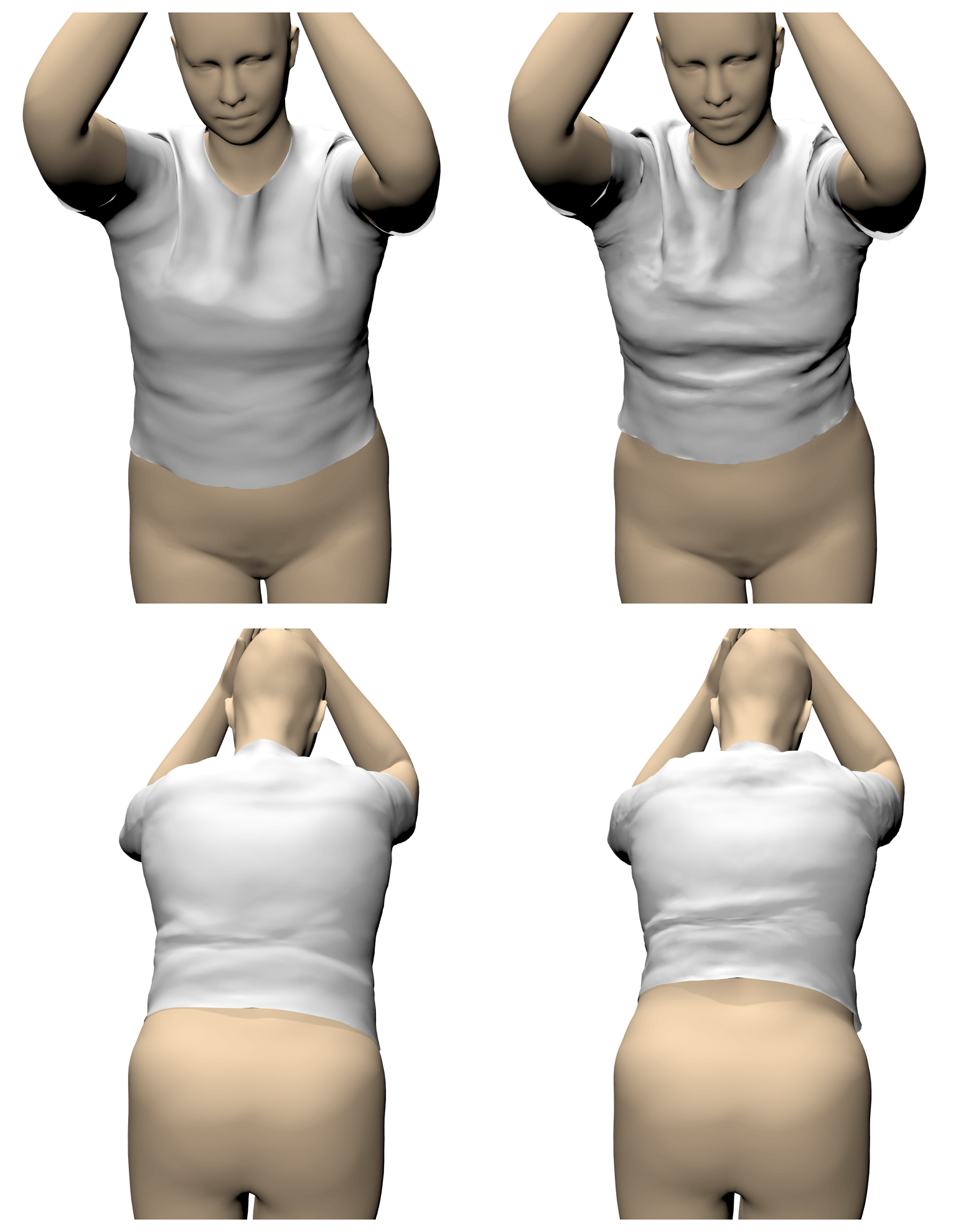}
	\caption{(left) Mixing the outputs of prototypes directly without decomposition smooths out the fine details. (right) The decomposition into high and low frequencies allows us to predict good garment fit with details. }
	\label{fig:freqdecomp}
\end{figure}
\begin{figure*}[ht]
	\centering
	\includegraphics[width=0.80\textwidth]{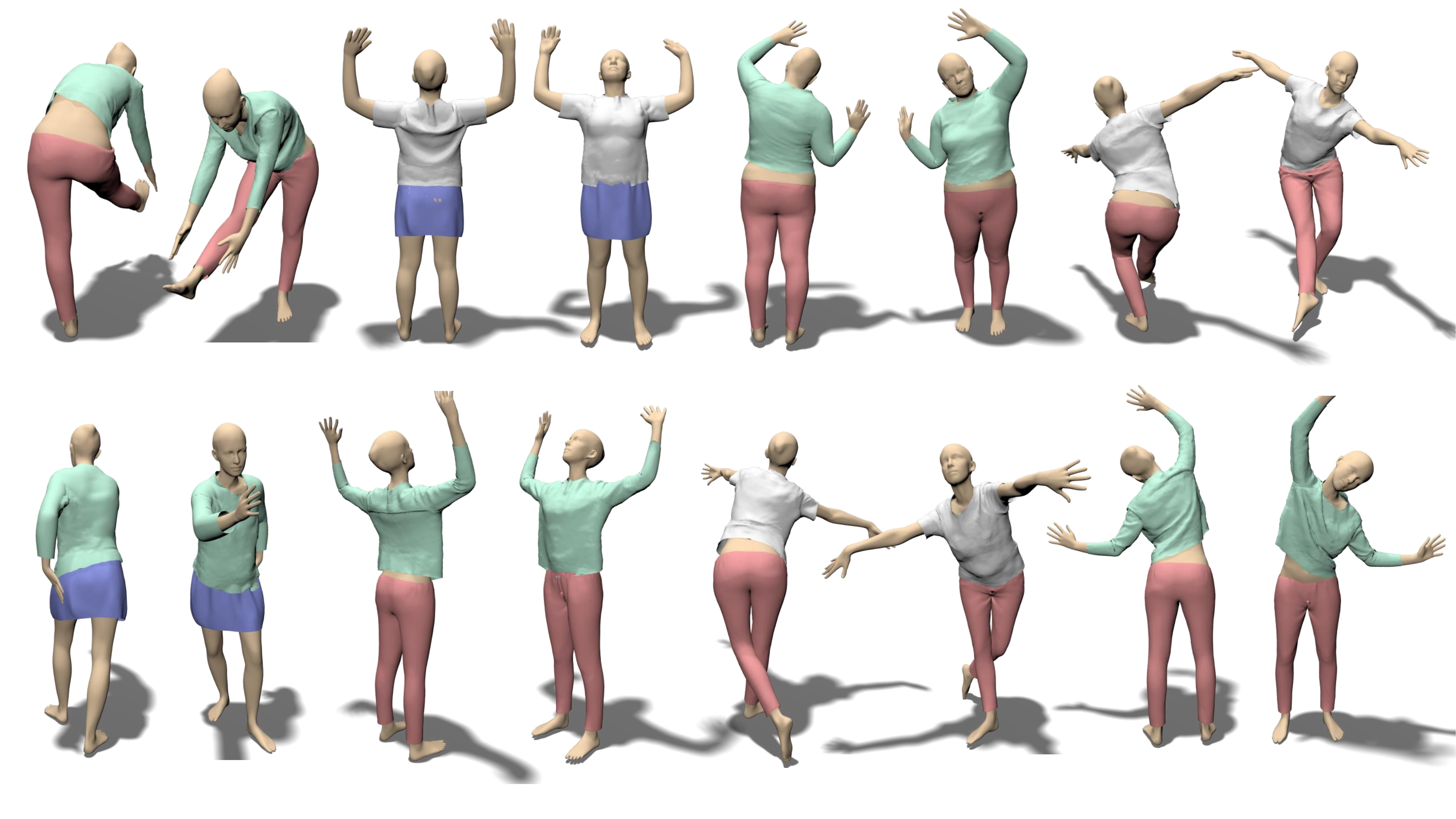}
	\includegraphics[width=0.19\textwidth]{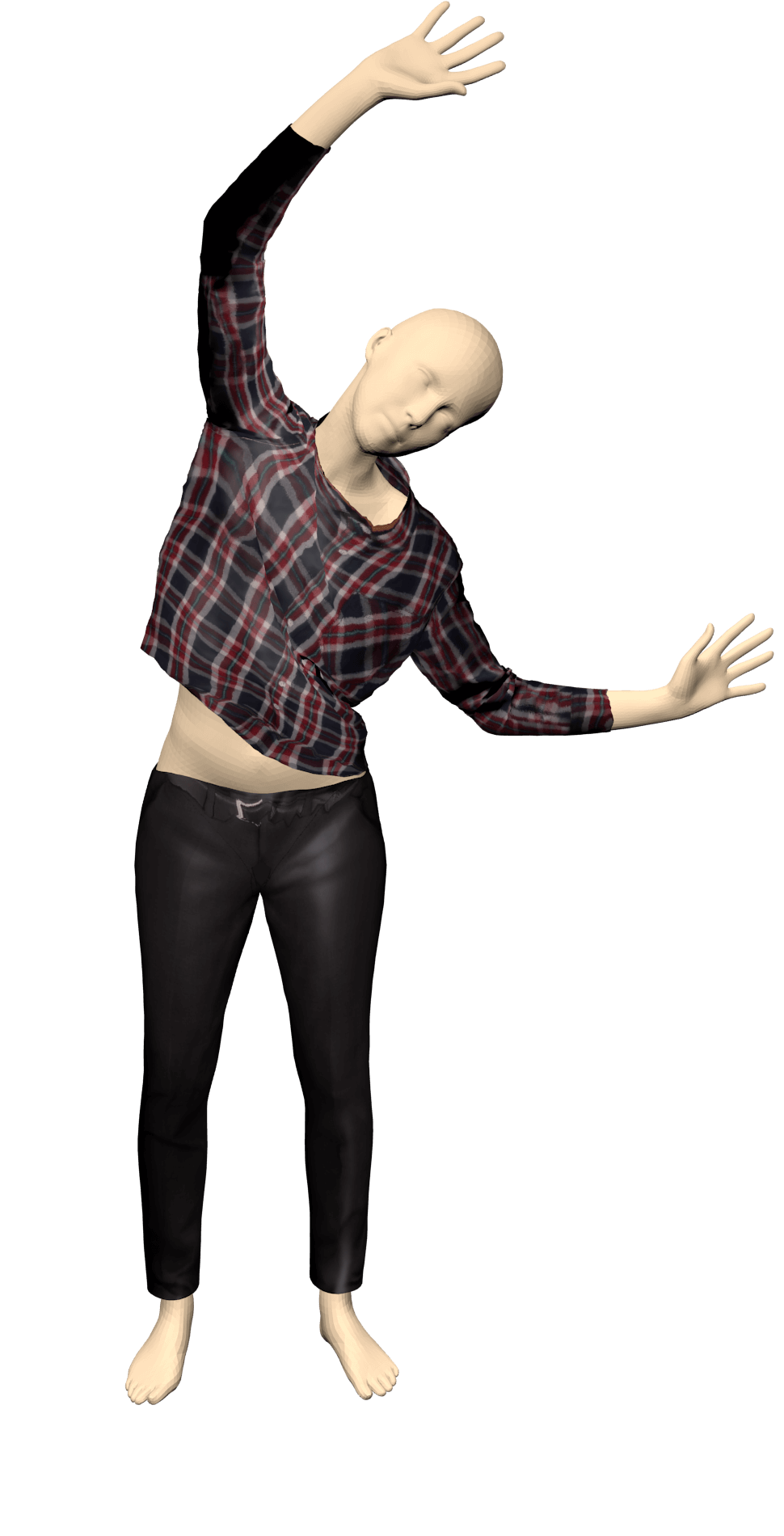}
	\caption{Left: Predictions of TailorNet for multiple garments and completely unseen poses. Right: TailorNet prediction with a real texture. }
	\label{fig:multigars}
\end{figure*}

We define our baseline $f_{BL}(\shape, \pose, \style) : \mathbb{R}^{|\pose|+|\shape|+|\style|} \rightarrow \mathbb{R}^{m\times3}$ implemented as a MLP to predict the displacements.
Table \ref{table:resultsblmixture} shows that our mixture model outperforms the baseline by a slight margin on 3 testing splits. 
\begin{table}[ht]
\centering
\begin{tabular}{c|c c|c c}
    \thickhline
    Split & Style-shape & Pose & Our & Our Mixture \\
    No. & set & set & Baseline & Model \\
     \hline
    2 & train & test & 10.6 & 10.2 \\
    3 & test & train & 11.7 & 11.4 \\
    4 & test & test & 11.6 & 11.4 \\
    \thickhline
\end{tabular}
\caption{Mean per vertex error in mm for 3 testing splits in Section~\ref{dataset:Simulation}. Our mixture performs slightly better than the baseline quantitatively, and significantly better qualitatively.}
\label{table:resultsblmixture}
\end{table}

Qualitatively, TailorNet outperforms the baseline and previous works. 
Figure \ref{fig:compareEG} shows the garment predicted by Santesteban \etal\cite{Casas19} and our mixture model.
Figure \ref{fig:compareBL} shows the qualitative difference between the output by TailorNet and our baseline trained on the same dataset.

To validate our choice to decompose high and low frequency details, we consider a mixture model where individual MLPs predict style-shape dependent displacements directly without decomposition. Figure \ref{fig:freqdecomp} shows that although it can approximate the overall shape well, it looses the fine wrinkles. TailorNet retains the fine details and provides intuitive control over high frequency details.

{\bf Sequences of AMASS~\cite{mahmood2019amass}}:  TailorNet generalizes (some poses shown in Fig.~\ref{fig:multigars}) despite being trained on completely different poses. 
Notably, the results are temporally coherent, despite not modelling the dynamics. See suppl. video for visualization.

\subsection{Multiple Garments}
To show the generalizability of TailorNet, we trained it separately for 3 more garments - Shirt, Pants and Skirt. Since skirt do not follow the topology of template body mesh, we attach a skirt template to the root joint of SMPL~\cite{Pons-Moll:Siggraph2017}. For each of these garments, we simulate the dataset for 9 shapes with a common style and trained the model. Figure~\ref{fig:multigars} shows the the detailed predictions by TailorNet. Since we use the base garments, which come from a \emph{real} digital wardrobe~\cite{bhatnagar2019mgn}, we can also transfer the textures from real scan on our predictions.


\subsection{Runtime Performance}
We implement TailorNet using PyTorch~\cite{pytorch}. On a gaming laptop with an NVIDIA GeForce GTX 1060 GPU and Intel i7 CPU, our approach runs from 1 to 2 ms per frame, which is 1000 times faster than PBS it is trained from. Just on CPU, it runs 100 times faster than PBS.


\section{Discussion and Conclusion}
\label{sec:conclusion}
TailorNet is the first data-driven clothing model of pose, shape and \emph{style}. 
The experiments show that a simple MLP approximates clothing deformations with an accuracy of $11$ mm, which is as good as more sophisticated graph-CNN, or displacements prediction in UV-space, 
but shares the same limitations with existing methods: when it is trained with different body shapes (and styles in our case) results are overly smooth and lack realism. 
To address this, we proposed a model which predicts low and high frequencies separately. 
Several experiments show that our narrow bandwidth mixture model to predict high-frequency preserves significantly more detail than existing models, despite having a much harder task, that is, modelling pose dependent deformation as a function of shape and style jointly.
 
In future work, we plan to fit the model to scans, images and videos, and will investigate refining it using video data in a self-supervised fashion, or using real 3D cloth captures~\cite{Pons-Moll:Siggraph2017}. We also plan to make our model sensitive to the physical properties of the cloth fabric and human soft-tissue~\cite{PonsMoll2015Dyna}.

Our TailorNet $1000$ times faster than PBS, allows easy control over style, shape and pose without manual editing. 
Furthermore, the model is differentiable, which is crucial for computer vision and learning applications. We will make TailorNet and our dataset publicly available, and will continuously add more garment categories and styles to make itl even more widely useable. 
TailorNet fills a key missing component in existing body models like SMPL and ADAM~\cite{Loper2015SMPL,joo2018total}: realistic clothing, 
which is necessary for animation of virtual humans, and to explain the complexities of dressed people in scans, images, and video.

\noindent\textbf{Acknowledgements.} This work is partly funded by the Deutsche Forschungsgemeinschaft (DFG, German Research Foundation) - 409792180 (Emmy Noether Programme, project: Real Virtual Humans) and Google Faculty Research Award. We thank Bharat Lal Bhatnagar and Garvita Tiwari for insightful discussions, and Dan Casas for providing us test results of \cite{Casas19}.

\newpage
{\small
	\bibliographystyle{ieee_fullname}
	\bibliography{CVPR20_final}
}
\end{document}